\documentclass{article}

\usepackage{microtype}
\usepackage{graphicx}
\usepackage{subfigure}
\usepackage{booktabs} % for professional tables

\usepackage{hyperref}

\usepackage[accepted]{icml2024}

\usepackage{amsmath}
\usepackage{amssymb}
\usepackage{mathtools}
\usepackage{amsthm}

\usepackage[capitalize,noabbrev]{cleveref}

\theoremstyle{plain}

\theoremstyle{definition}

\theoremstyle{remark}

\usepackage[textsize=tiny]{todonotes}

\icmltitlerunning{All Roads Lead to Rome?}

\begin{document}

\twocolumn[
%\icmltitle{\textit{All Roads Lead to Rome?} Exploring Representational Similarities Between Latent Spaces of Generative Image Models}
\icmltitle{All Roads Lead to Rome? Investigating Representational \\Similarities in Generative Image Models}




\begin{icmlauthorlist}
\icmlauthor{Charumathi Badrinath}{Harvard}
\icmlauthor{Usha Bhalla}{Harvard}
\icmlauthor{Alex Oesterling}{Harvard}
\icmlauthor{Suraj Srinivas}{Harvard}
\icmlauthor{Himabindu Lakkaraju}{Harvard}
\end{icmlauthorlist}

\icmlaffiliation{Harvard}{Harvard University}

\icmlcorrespondingauthor{Charumathi Badrinath}{charumathi.badrinath@gmail.com}

\icmlkeywords{Machine Learning, ICML}

\vskip 0.3in
]



\printAffiliationsAndNotice{} % otherwise use the standard text.

\begin{abstract}
Do different generative image models secretly learn similar underlying representations? We investigate this by measuring the latent space similarity of four different models: VAEs, GANs, Normalizing Flows (NFs), and Diffusion Models (DMs). Our methodology involves training linear maps between frozen latent spaces to ``stitch" arbitrary pairs of encoders and decoders and measuring output-based and probe-based metrics on the resulting ``stitched'' models. Our main findings are that linear maps between the latent spaces of performant models preserve most visual information even when latent space sizes differ; for CelebA models, \textit{gender} is the most similarly represented probe-able attribute. Finally we show on a Normalizing Flow that latent space representations converge early in training.
\end{abstract}

\section{Introduction}
Recent literature on deep networks has suggested that all models are converging to the same representation of ``reality". Huh et al. (\citeyear{huh2024platonic}) show that the embedding spaces of performant vision and language models learn similar notions of ``distance" between datapoints of the appropriate modalities. Bansal et al. (\citeyear{bansal2021revisiting}) ``stitch" together layers of pairs of powerful vision networks via a simple map and show that the combined network has minimal loss in performance. 

Asperti and Tonelli (\citeyear{asperti2022map}) investigate whether a similar conclusion holds for \textit{latent spaces} of generative image models by extending the model stitching framework. They train a \textit{linear} map between the latent spaces of a Variational Autoencoder (VAE) and Generative Adversarial Network (GAN) and show that this stitched VAE-encoder-GAN-decoder model is able to produce reconstructions of images that look similar to the originals and have a low RMSE.

However, despite previous work, it remains unclear how well these conclusions hold for other families of generative image models, especially those with vastly different latent sizes. Further, it is not clear which \textit{specific} visual attributes are represented similarly by models and which aren't. In our work, we address these drawbacks by (1) extending the study to include NFs and DMs, (2) including probe-based metrics to capture semantics of representations. On CelebA we find that powerful models can be stitched via their latent spaces to reconstruct an image with little loss in quality. In addition to color, pose, and lighting, attributes related to \textit{gender} are most similarly represented. Finally, we explore the question of how quickly the local minimum corresponding to this shared latent space is found, by seeing how early in training latent space probe accuracy plateaus. On an NF, we find that this happens less than 20\% of the way into training. 

Our overall contributions are:
\vspace{-10pt}
\begin{enumerate}
\itemsep0em 
    \item In \S \ref{sec:metrics}, we propose two sets of metrics (reconstruction-based, and probe-based metrics) grounded in a model ``stitching"-based procedure to measure the similarity of generative image model latent spaces. 
    \item In \S \ref{sec:results}, we perform extensive experiments across four generative image model classes trained on CelebA (VAEs, GANs, NFs and DMs), and find significant similarity among their latent spaces on our metrics, and especially in their representation of gender. On the Normalizing Flow model, we find that this common latent space structure is learned early in training.
\end{enumerate}

\section{Related Works}
\subsection{Exploring Latent Spaces}
\textbf{Discovering latent directions}. Several approaches have been proposed to discover vectors corresponding to high-level image concepts (e.g. hair color) in the GAN latent space. Supervised methods include subtracting the average latent vectors corresponding to images with and without a binary attribute \cite{larsen2016autoencoding} and finding hyperplanes in the latent space separating images with different attribute values \cite{shen2020interfacegan}. One unsupervised method uses PCA on the GAN latent space to discover the most salient edit directions \cite{härkönen2020ganspace}. These techniques have been applied to find edit directions in the latent spaces of VAEs \cite{white2016sampling} and NFs \cite{kingma2018glow} as well. Our work \textit{compares} the structure of different models' latent spaces and is thus orthogonal to the exploration of latent spaces in isolation.

\subsection{Comparing Model Representations}
\textbf{Metrics.} Canonical correlation analysis \cite{morcos2018cca} and centered kernel alignment \cite{kornblith2019cka} are metrics that have been used to measure representational similarity between layers of different models. Another work compared distances between sets of datapoints in different models' embedding spaces \cite{huh2024platonic}. 

\textbf{Model Stitching.} One branch of work compares representations via ``stitching" where a simple map is learned between layers of two different models \cite{lenc2015understanding}. One work stitched together neural networks of different strengths at an intermediate layer, concluding that good models tended to converge on similar representations and that stitching strong models to weak models improved their performance \cite{bansal2021revisiting}. Work on stitching the latent spaces of generative models has also been explored. One paper showed that it is possible to construct a \textit{linear} map between the latent space of image encoders and text decoders to obtain text describing an image, demonstrating that models learn similar representations of multi-modal data \cite{merullo2023linearly}. 

The work most closely related to ours showed that it is possible to linearly stitch the latent spaces of a VAE and GAN and reconstruct an image with little loss in quality \cite{asperti2022map}. Our work extends this idea to more model architectures, performing a thorough pixel-space and latent-space analysis to characterize what types of models learn similar representations and what specific concepts they represent similarly.

\section{Methodology}
\subsection{Models and Datasets}
\begin{table}[h]
    \centering
    \begin{tabular}{|c||c|c|c|c|c|}
        \hline
        Model & GAN & VAE & VQVAE & NF & DM \\
        \hline
        Latent Dim & 512 & 512 & 768 & 12288 & 12288 \\
        \hline
    \end{tabular}
    \caption{Models and their latent space sizes. See Appendix \ref{app:bgd} for background and Appendix \ref{app:models} for more details on the model architectures and implementations used.}
    \label{tab:model-architectures}
    \vspace{-3pt}
\end{table}

We experiment on 5 generative image models trained on CelebA \cite{celeba} -- a dataset of celebrity faces annotated with 40 binary attributes. Models (summarized in Table \ref{tab:model-architectures}) output images of size 64 $\times$ 64 with slightly different crops.

We use a GAN-generated dataset \textit{CelebA-Synthetic} to train linear maps to and from the GAN latent space, since GANs lack an encoder. The ``latent space" of the DM is taken to be the noised image after running the forward process using the DDIM scheduler with 50 timesteps \cite{song2020denoising}. As a baseline, we use a ``random" encoder which maps images deterministically to arbitrary 512-dimensional standard Gaussian vectors.

% Unlike those of other models, this latent space isn't a true representation space as noising does not induce a representation in the same way as convolutions or compression.  

\subsection{Proposed Metrics}
\label{sec:metrics}
Model latent spaces are first stitched via linear maps (see Appendix \ref{app:lin-maps}).\\
\\
\textbf{Reconstruction-Based Metrics.} We qualitatively assess the image reconstructions of stitched models. We also compute the latent space MSE of the mapping, pixel-space RMSE of the reconstructed image, LPIPS score \cite{zhang2018lpips} which measures the perceptual similarity between a pair of images, and the FID \cite{heusel2018fid} which measures how similar the distribution of reconstructed images is to the distribution of target images. Lower is better for all metrics.\\
\textbf{Probe-Based Metrics.} We train linear probes (see Appendix \ref{app:lin-probes}) on the latent spaces of encoder-decoder models to detect the presence of binary attributes and report test set accuracy. We then measure the percentage of times probe predictions on latents encoded by the model whose latent space it was trained on and those mapped from other latent spaces match (higher is better). We also measure the percent change in probe accuracy on these sets of latents (near 0 is better).

\section{Results}
\label{sec:results}
\subsection{Assessing Stitched Model Reconstructions}
\label{sec:assessing}

We stitch the latent spaces of a VAE, VQVAE, NF, DM and GAN pairwise and compute the reconstruction-based metrics described in Section \ref{sec:metrics}.

\begin{figure}[H]
\centering
 \raisebox{2.5\height}{
\includegraphics[width=0.11\columnwidth]{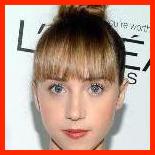}
}
\includegraphics[width=0.65\columnwidth]{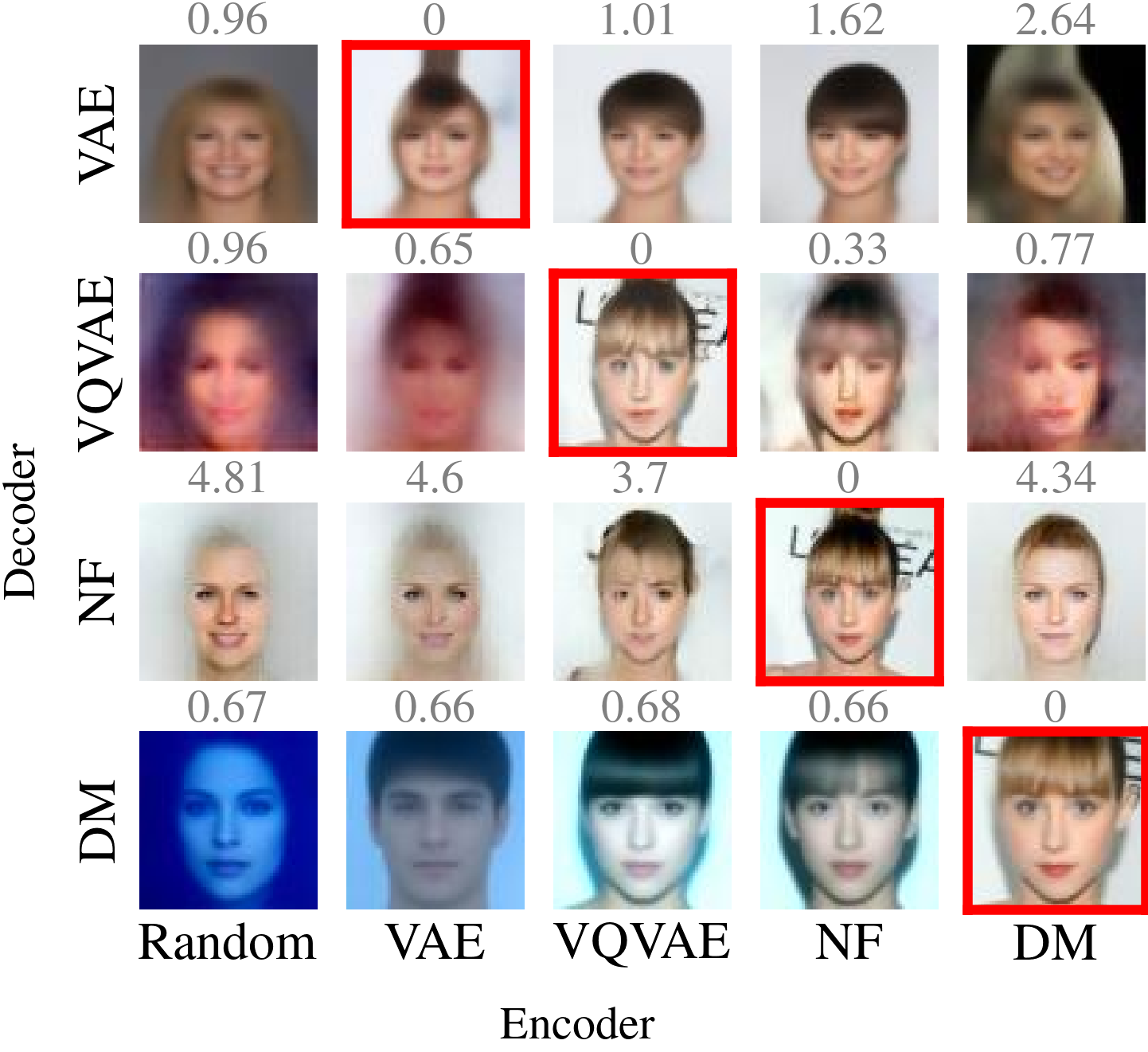}
\vskip -0.1in
\caption{Reconstruction of a CelebA image (left) using stitched models. Latent space MSEs of mapped latents are displayed above each image. The stitched models yielding the best reconstructions use the VQVAE and NF encoders.}
\label{fig:celeba-mapped}
\end{figure}

From Figure \ref{fig:celeba-mapped} we see that stitching the NF and VQVAE encoders to any model $\mathbb{X}$'s decoder yields reconstructions resembling those generated by $\mathbb{X}$ (boxed in red) and obtain relatively low latent MSE, RMSE, LPIPS and FID scores (Figure \ref{fig:celeba-metrics}). Additional examples are in Appendix \ref{app:examples}.

\begin{figure}[H]
\centering
\includegraphics[width=0.8\columnwidth]{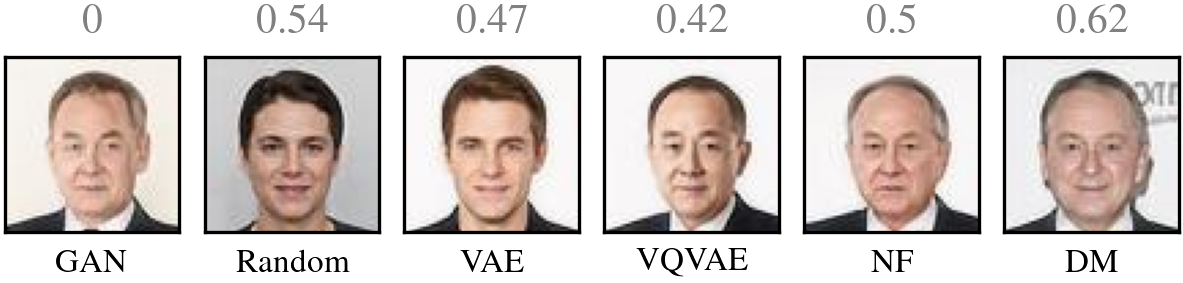}
\vskip -0.1in
\caption{Reconstruction of a CelebA-Synthetic image using various encoders stitched to the GAN decoder. The leftmost image is the ground truth. Stitched models using the VQVAE and NF encoders yield the closest reconstructions.}
\label{fig:celeba-mapped-gen}
\end{figure}

Conversely, the DM and VAE encoders barely outperform the random encoder in terms of stitched reconstruction quality. Poor VAE performance can be attributed to significant information loss during encoding which is reflected in lossy reconstructions even with its own decoder. For the DM, we note that the ``latent vectors" we use to construct our mappings are heavily noised versions of the image rather than true encodings. The consequence of this is poor linear probe performance (Figure \ref{fig:post-map-accuracies}). Latent space maps pick up on remaining signal, which is sometimes enough to map to an area of the decoder model's latent space producing images in the visual vicinity of an target image (Figure \ref{fig:celeba-mapped-gen}). The spatial nature of DM ``latents" (the``representation" of a feature is a function of the pixel values in the region of the image that the feature manifests) makes the ability to stitch another model's encoder to its decoder \textit{at all} unexpected. However, it explains why models stitched to the DM decoder often produce reconstructions preserving the geometry of the target image but not the colors. 

Finally, Figure \ref{fig:celeba-mapped-gen} shows reconstructions of a test image from CelebA-Synthetic obtained by stitching various models' encoders to the GAN decoder. Due to the strength of the GAN decoder, all reconstructions resemble a face. Furthermore, mapping to the vicinity of the true GAN latent is sufficient to produce a reconstruction preserving most characteristics of the target image (e.g. age, hair and skin color, expression) supporting results from literature \cite{asperti2022map}. 

\begin{figure}[ht]
\centering
\includegraphics[width=0.45\columnwidth]{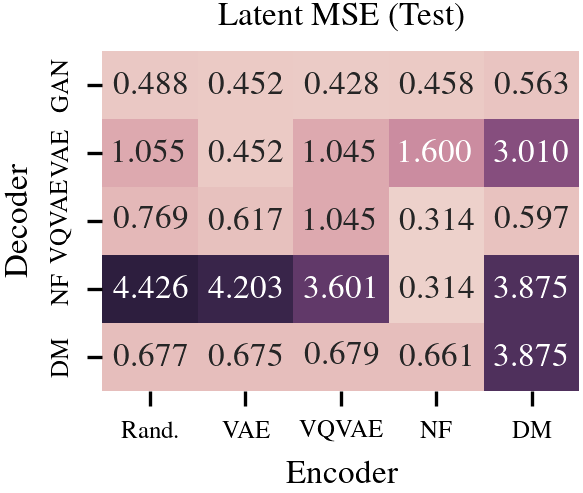}
\includegraphics[width=0.45\columnwidth]{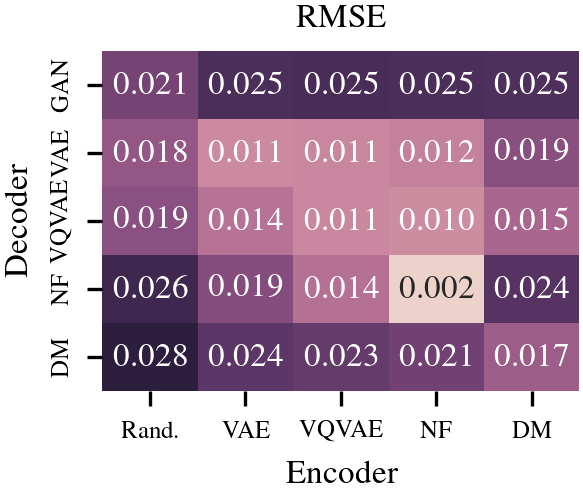}
\includegraphics[width=0.45\columnwidth]{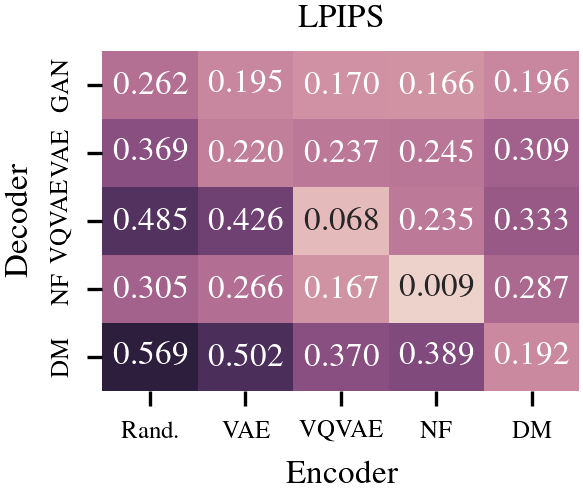}
\includegraphics[width=0.45\columnwidth]{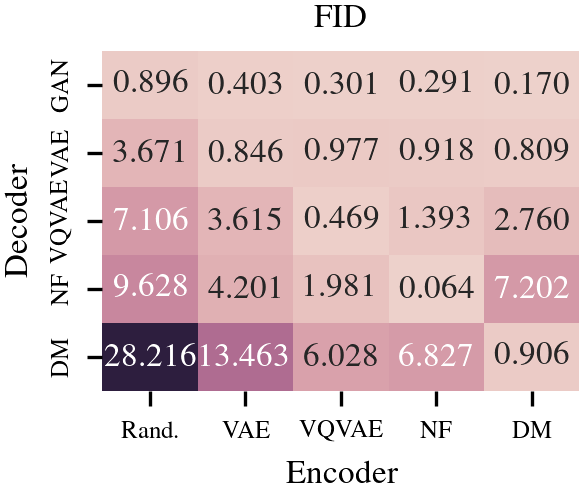}
\vskip -0.1in
\caption{Heatmaps of latent-space MSE, pixel-space RMSE, LPIPS and FID for images reconstructed by stitched models. We see relatively low values of these metrics for models using the NF and VQVAE encoders, corroborating the results of Figure \ref{fig:celeba-mapped}.}
\label{fig:celeba-metrics}
\end{figure}

\subsection{Identifying Similarly Represented Attributes}
\label{sec:sim-rep}

\begin{figure}[h]
\centering
\includegraphics[width=0.75\columnwidth]{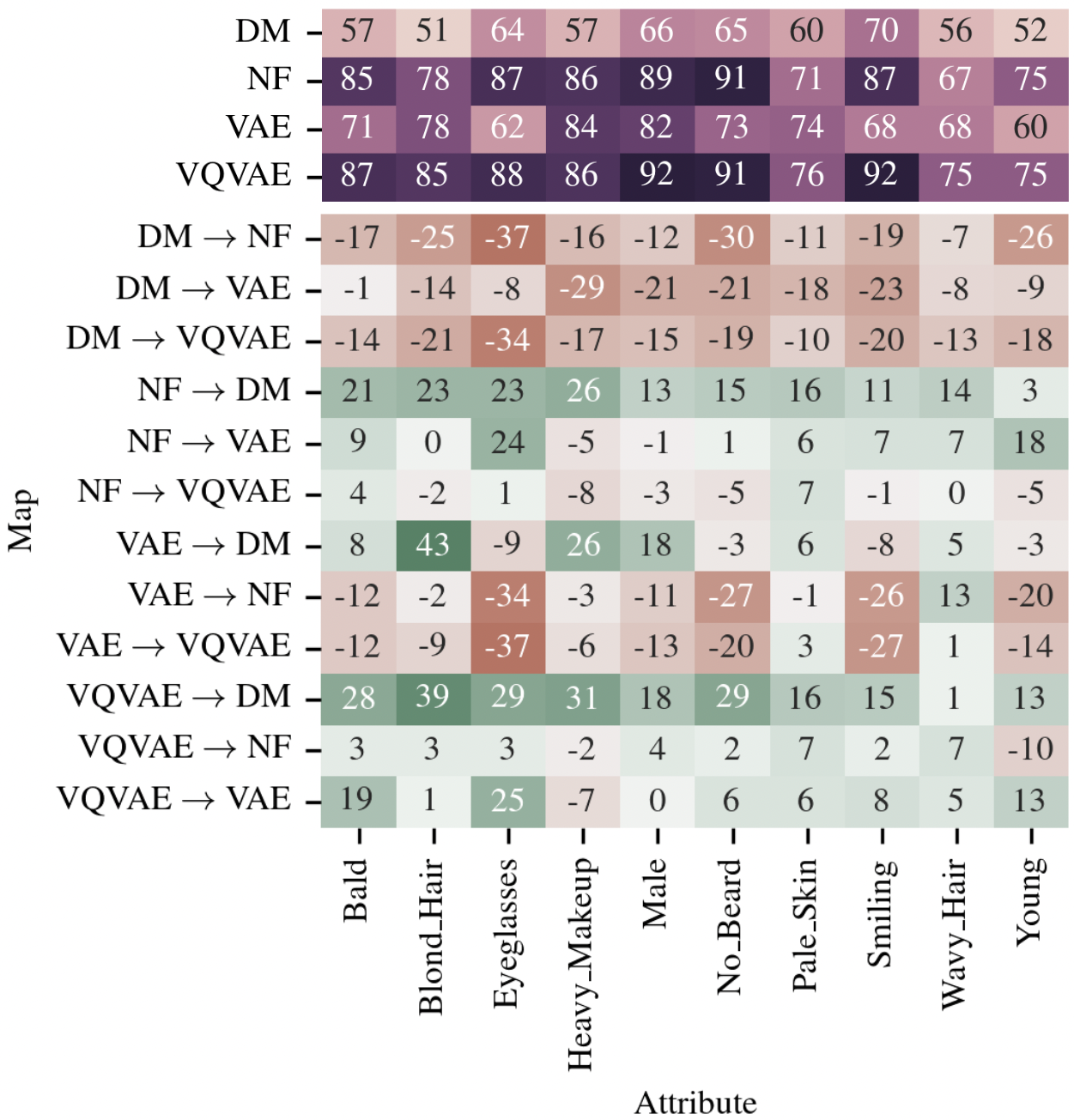}
\vskip -0.1in
\caption{The top 4 rows show latent space linear probe accuracy on various binary attributes. The NF and VQVAE have the most linearly separable latent spaces. The remaining rows show change in accuracy of probe trained on model $\mathbb{X}$'s latent space when making predictions on latents encoded by $\mathbb{X}$ and latents mapped from $\mathbb{X}' \rightarrow \mathbb{X}$. Probe accuracy increases on latents mapped from a more linearly separable to less linearly separable latent space.}
\label{fig:post-map-accuracies}
\vskip -0.1in
\end{figure}

We stitch the latent spaces of a VAE, VQVAE, NF, DM and GAN pairwise and compute the probe-based metrics described in Section \ref{sec:metrics} on 10 interesting CelebA attributes. Results for all 40 attributes are in Appendix \ref{app:probes}.  

\begin{figure}[h]
\centering
\centerline{\includegraphics[width=0.75\columnwidth]{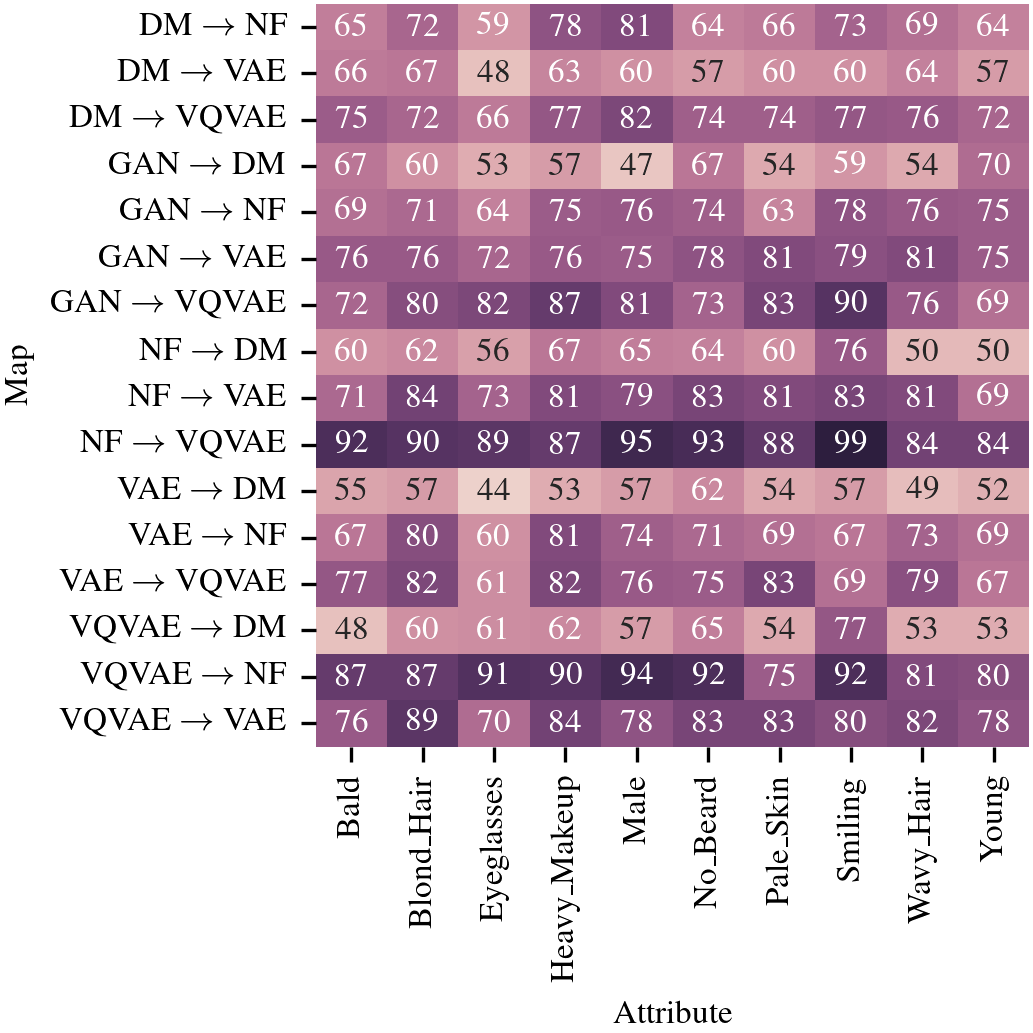}}
\vskip -0.1in
\caption{Percentage of times a probe trained on model $\mathbb{X}$'s latent space produces the same prediction on latents encoded by $\mathbb{X}$ and latents mapped from $\mathbb{X}' \rightarrow \mathbb{X}$. Attributes correlated with gender are represented similarly by nearly every model.}
\label{fig:post-map-matches}
\vskip -0.2in
\end{figure}

From Figure \ref{fig:post-map-accuracies} we see that linear probes for the gender-related attributes \texttt{Male} and \texttt{Heavy Makeup} achieve \>80\% accuracy on the NF, VAE and VQVAE latent spaces. In Figure \ref{fig:post-map-matches} we see that NF probes produce similar predictions on latents produced by its own decoder and latents encoded and mapped from the VQVAE latent space (and vice versa), in agreement with the results from Section \ref{sec:assessing}. 

Figure \ref{fig:post-map-matches} also shows a relatively high percentage of matches between probes on NF, VAE, and VQVAE latents, and their counterparts mapped from the GAN latent space. Since the GAN latent space is also known to be linearly separable \cite{larsen2016autoencoding}, we conclude that latent space ``stitchability" of two models depends on the linearity of both models' latent representations.

We hypothesize that attributes that are represented similarly in all models' latent spaces are either uniformly represented in the training dataset (e.g. \texttt{Smiling}), or explain much of the variance in an image (e.g. \texttt{Male} and related attributes). Interestingly, post-map probe match percentages are high for gender-related attributes, particularly \texttt{Heavy\_Makeup} and \texttt{Male}, even for for maps originating from the non-linear DM latent space indicating that signal corresponding to these attributes is least likely to be lost in the noising process.

Finally, we note in Figure \ref{fig:post-map-accuracies} that probes trained on less linearly separable latent spaces are often \textit{more} accurate when applied on latents mapped from a more linearly separable latent space. This explains the phenomenon of the corresponding reconstructions having more ``extreme" versions of the attributes present in the target image. We see an example of this in Figure \ref{fig:celeba-mapped} where the reconstructions produced by the NF and VQVAE stitched to the DM have very distinct, block-like bangs. 

\subsection{Observing Latent Space Structure Over Training}
\begin{figure}[h]
%\vskip 0.2in
\centering
\includegraphics[width=0.9\columnwidth]{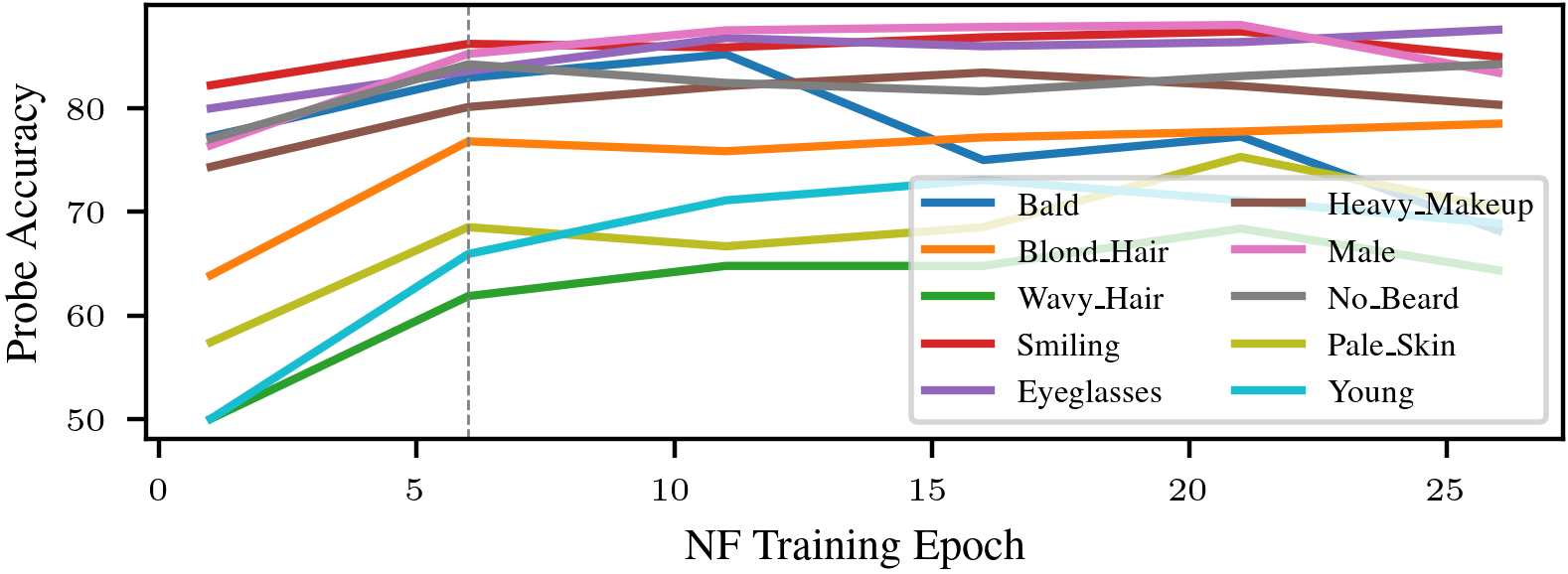}
%\vskip -0.1in
\caption{Test set accuracy of linear probes on NF latent space frozen at increasing training epochs. Probe accuracy on most attributes seems to plateau after epoch 6.}
\label{fig:nf-accuracies}
\end{figure}

\begin{figure}[h]
\centering
\centerline{\includegraphics[width=0.7\columnwidth]{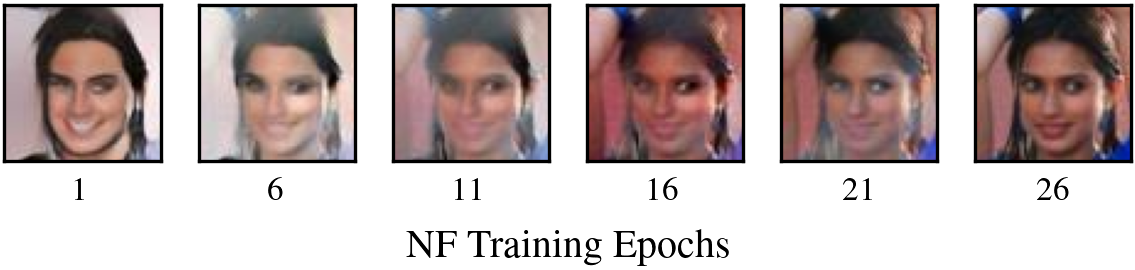}}
%\vskip -2.0
\caption{Reconstructions generated by encoding an image with a partially trained NF trained and decoding with a fully trained NF. Reconstruction quality improves with encoder training.}
\label{fig:nf-decoded}
\vspace{-15pt}
\end{figure}

We train the Glow NF \cite{kingma2018glow} on CelebA for 26 epochs checkpointing every 5 epochs. We train linear probes for 10 interesting attributes on the frozen latent spaces and measure how their accuracy changes as model training progresses.

Figure \ref{fig:nf-accuracies} shows that probe accuracy begins to plateau for all attributes by epoch 6 of 26. Attributes correlated with gender as well as \texttt{Smiling} (the similarly represented attributes identified in Section \ref{sec:sim-rep}) change minimally in their representations after epoch 1 of 26. Figure \ref{fig:nf-decoded} shows that there is slight improvement in images reconstructed from latents encoded by increasingly trained NF models, indicating that the latent space still changes over time, albeit minimally after epoch 11. Thus, we hypothesize that a ``universal" representation is natural for models to learn and that later training epochs improve the decoder given a somewhat fixed latent representation.

\section{Discussion and Future Work}
This work advances the hypothesis that as (unconditional) generative image models become more expressive, their latent space representations (given that they are meaningful) become more similar to one another. We find that this shared representation is learned early on in model training, indicating that it is quite ``natural" in describing the data. This suggests that the latent spaces of increasingly expressive models may be converging to an ``optimal" data representation in the limit, which has deep implications for how we construct world models.

Similarity of latent space representations also means that edit directions, unsafe regions, or areas of bias found in the latent space of one model, can be mapped onto any other model via a simple linear map. It also allows for more powerful image editing by stitching models with strong encoders to models with strong decoders via their latent spaces.

Future work can investigate whether similar results hold for models trained on datasets containing images from multiple classes with more variance, and can probe similarity of representation of non-binary attributes. We expect to see similar results given that the models used are able to produce realistic samples from the data distribution.
% There are several avenues for future work beyond the replication of our experiments on a wider range of models and datasets. First, we can investigate whether convergence to similar representations is a characteristic of latent space models capable of generation, or whether it extends to auto-encoders or dimensionality reduction algorithms as well. 
We also wish to extend our analysis to pipelines composed of \textit{multiple} foundation models, for example, the latent diffusion model \cite{ldm} which trains a DM on the VQVAE latent space. Since it is the \textit{first} foundation model in this pipeline that learns a latent representation of the data, we expect to see similar results to this paper. Finally, work has shown that it is possible to find a linearly separable latent space for DMs \cite{dm-latentspace}; we can validate that this latent space is similarly structured to those studied in our paper.

% \section*{Software and Data}

% If a paper is accepted, we strongly encourage the publication of software and data with the
% camera-ready version of the paper whenever appropriate. This can be
% done by including a URL in the camera-ready copy. However, \textbf{do not}
% include URLs that reveal your institution or identity in your
% submission for review. Instead, provide an anonymous URL or upload
% the material as ``Supplementary Material'' into the OpenReview reviewing
% system. Note that reviewers are not required to look at this material
% when writing their review.

\section*{Acknowledgements}
We would like to thank the GRaM workshop's ELLIS Mobility Grant as well as the SPIGM workshop for supporting CB's travel and attendance at ICML. 

% \section*{Impact Statement}
% This paper presents work whose goal is to advance the field of 
% Machine Learning. There are many potential societal consequences 
% of our work, none which we feel must be specifically highlighted here.


\bibliography{example_paper}
\bibliographystyle{icml2024}

%%%%%%%%%%%%%%%%%%%%%%%%%%%%%%%%%%%%%%%%%%%%%%%%%%%%%%%%%%%%%%%%%%%%%%%%%%%%%%%
%%%%%%%%%%%%%%%%%%%%%%%%%%%%%%%%%%%%%%%%%%%%%%%%%%%%%%%%%%%%%%%%%%%%%%%%%%%%%%%
% APPENDIX
%%%%%%%%%%%%%%%%%%%%%%%%%%%%%%%%%%%%%%%%%%%%%%%%%%%%%%%%%%%%%%%%%%%%%%%%%%%%%%%
%%%%%%%%%%%%%%%%%%%%%%%%%%%%%%%%%%%%%%%%%%%%%%%%%%%%%%%%%%%%%%%%%%%%%%%%%%%%%%%
\newpage
\appendix
\onecolumn
\section{Appendix}
\subsection{Source Code}
All code for this paper is available in the following GitHub repository: \href{https://github.com/charumathib/thesis-latent-spaces}{https://github.com/charumathib/thesis-latent-spaces}.
\subsection{Background on Models}
\label{app:bgd}
All generative image models learn to ``sample" images following their the distribution of their training set. However, the four main families of models --  Variational Autoencoders (VAEs), Generative Adversarial Networks (GANs), Normalizing Flows (NFs) and Diffusion Models (DMs) -- differ drastically in how they do so.

A \textbf{VAE} \cite{vaes} consists of an encoder network mapping images to latent vectors of lower dimension, and a decoder neural network that maps latent vectors to images with the objective of minimizing image reconstruction error. A KL-Divergence term is added to the objective to encourage a marginally Gaussian latent space distribution.

A \textbf{GAN} \cite{gans} is comprised of a generator network (decoder) that learns to generate images from marginally Gaussian latent samples and a discriminator network whose objective is to distinguish images from the training set from generated images. Both networks improve by playing a minimax game, ultimately yielding a generator capable of producing realistic images.

An \textbf{NF} \cite{nfs} samples a latent variable from a simple distribution and repeatedly applies a series of invertible functions to this variable until it is transformed into an image. The marginal likelihood of each datapoint can be derived via the change of variables theorem for distributions and the objective is the negative data log likelihood.

Finally, a \textbf{DM} \cite{ho2020denoising} progressively adds Gaussian noise to an image producing a sequence noisy samples that become isotropic to a Gaussian in the limit (``forward process"). It then learns how to produce samples from the ``reverse process" to gradually transform pure noise into an image. DMs dont have an obvious latent space in the same way as VAEs, GANs and NFs since marginally Gaussian ``latents" are created via progressive noising rather then by a transformation that would force a representation to be learned. 

\subsection{Models Used}
\label{app:models}
The pre-trained GAN we use is from the Diffusion-GAN codebase \cite{wang2022diffusiongan} and follows the StyleGAN2-ADA architecture \cite{karras2020training}. The pre-trained VAE is from the DiffuseVAE codebase \cite{diffusevae}. The pre-trained VQVAE \cite{oord2018neural} from the Hugging Face \texttt{diffusers} library \cite{diffusers}. The NF implementation is from an independent source \cite{nf_code_celeba} follows the Glow architecture \cite{kingma2018glow}. The DM is from the DDIM codebase \cite{song2020denoising}. 

\subsection{Training Latent Space Linear Maps}
\label{app:lin-maps}
Many of our experiments rely on the construction of linear maps between model latent spaces. For each pair of encoder-decoder models (VAE, VQVAE, NF, DM), we encode the first 9000 images from the CelebA test split into their respective latent spaces and use a closed-form linear regression solver from \texttt{sklearn} to predict one set of latents from the other \cite{scikit-learn}. To prevent overfitting in maps originating from the NF and DM latent spaces, we use ridge regression with manually tuned values of $\alpha$ as summarized in Table \ref{tab:alpha}.

\begin{table}[ht]
    \centering
    \begin{tabular}{|c|c|}
       \hline 
       Map  &  $\alpha$\\
       \hline
       DM $\rightarrow$ GAN  & 2000\\
       DM $\rightarrow$ VAE & 100\\
       DM $\rightarrow$ VQVAE & 5000\\
       DM $\rightarrow$ NF & 5000\\
       NF $\rightarrow$ GAN & 50000\\
       NF $\rightarrow$ VAE & 5000\\
       NF $\rightarrow$ VQVAE & 50000\\
       NF $\rightarrow$ DM & 50000\\
       \hline
    \end{tabular}
    \caption{Regularization parameters used in Ridge regression for each linear map. Maps that are not listed in this table were not regularized.}
    \label{tab:alpha}
\end{table}

Our training procedure differs slightly for maps to and from the GAN latent space. We use the ``style space" as our latent space (a layer connected to the base latent space by an 8 layer MLP) \cite{karras2019stylegan}. We use the first 9000 images in the CelebA-Synthetic dataset to train maps in the way described earlier; the latent used by the GAN to generate each image is treated as the image's ``encoding" in the GAN latent space. All maps are tested on a holdout set of 100 images from either CelebA or CelebA-Synthetic as appropriate.

\begin{figure}[ht]
\centering
\includegraphics[width=0.24\columnwidth]{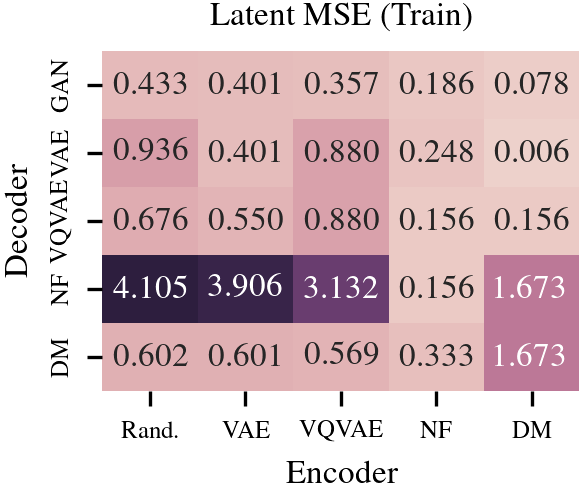}
\includegraphics[width=0.24\columnwidth]{figs/latent_mse_test_celeba.png}
\caption{Latent space map MSE on train (left) and test (right) sets.}
\vskip -0.2in
\end{figure}

\subsection{Training Latent Space Linear Probes}
\label{app:lin-probes}
We train linear probes to predict the value of each of the 40 binary attributes (e.g. \texttt{Male}) CelebA images are annotated with. To construct a balanced training set for an attribute's probe, we first compute the number of images with and without that attribute in the first 9000 CelebA images. 80\% of the minimum of these two numbers is how many positive and negative examples are included. Probes are trained using lasso regression in \texttt{sklearn} where $\alpha = 0.005$ for probes on the VAE and VQVAE latent spaces, 0.02 for probes on the DM latent space and 0.1 for probes on the NF latent space. \cite{scikit-learn}. We use a holdout set of 200 latents (100 with attribute, 100 without attribute) to evaluate probe performance.

\subsection{Full Probe Results}
\label{app:probes}

\begin{figure}[ht]
\centering
\includegraphics[width=\columnwidth]{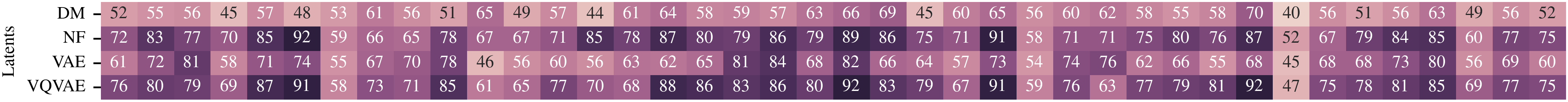}
\caption{Full version of Figure \ref{fig:post-map-accuracies} (top) with results for all 40 attributes.}
\end{figure}

\begin{figure}[ht]
\centering
\includegraphics[width=\columnwidth]{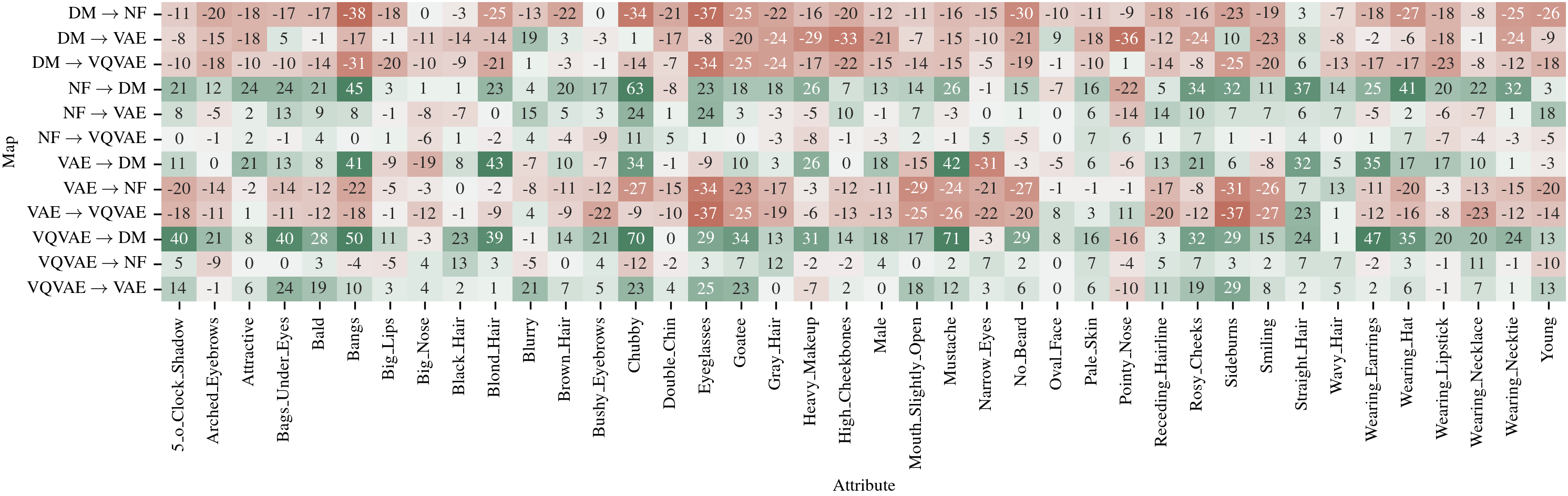}
\caption{Full version of Figure \ref{fig:post-map-accuracies} (bottom) with results for all 40 attributes.}
\end{figure}

\begin{figure}[ht]
\centering
\includegraphics[width=\columnwidth]{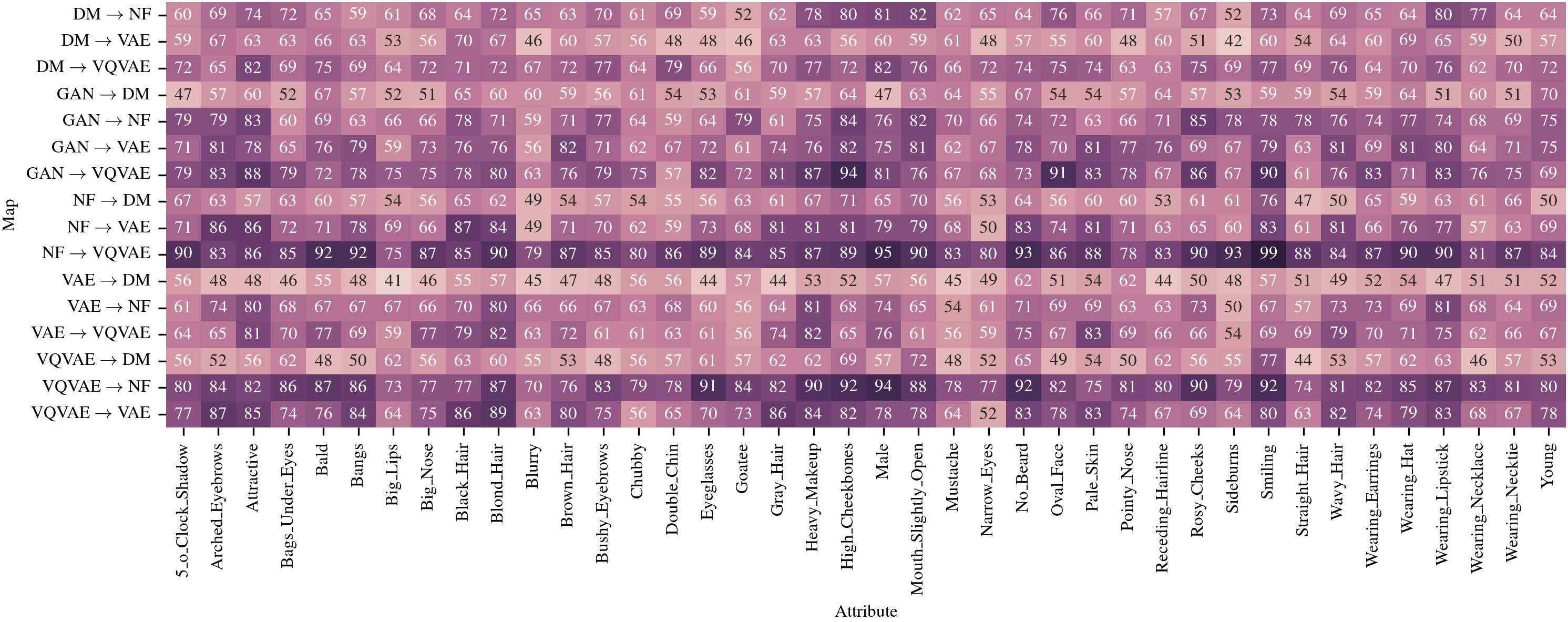}
\caption{Full version of Figure \ref{fig:post-map-matches} with results for all 40 attributes.}
\end{figure}

\subsection{Additional Examples of Stitched Model Reconstructions}
\label{app:examples}

\begin{figure}[H]
\centering
 \raisebox{2\height}{
\includegraphics[width=0.07\columnwidth]{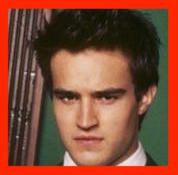}
}
\includegraphics[width=0.35\columnwidth]{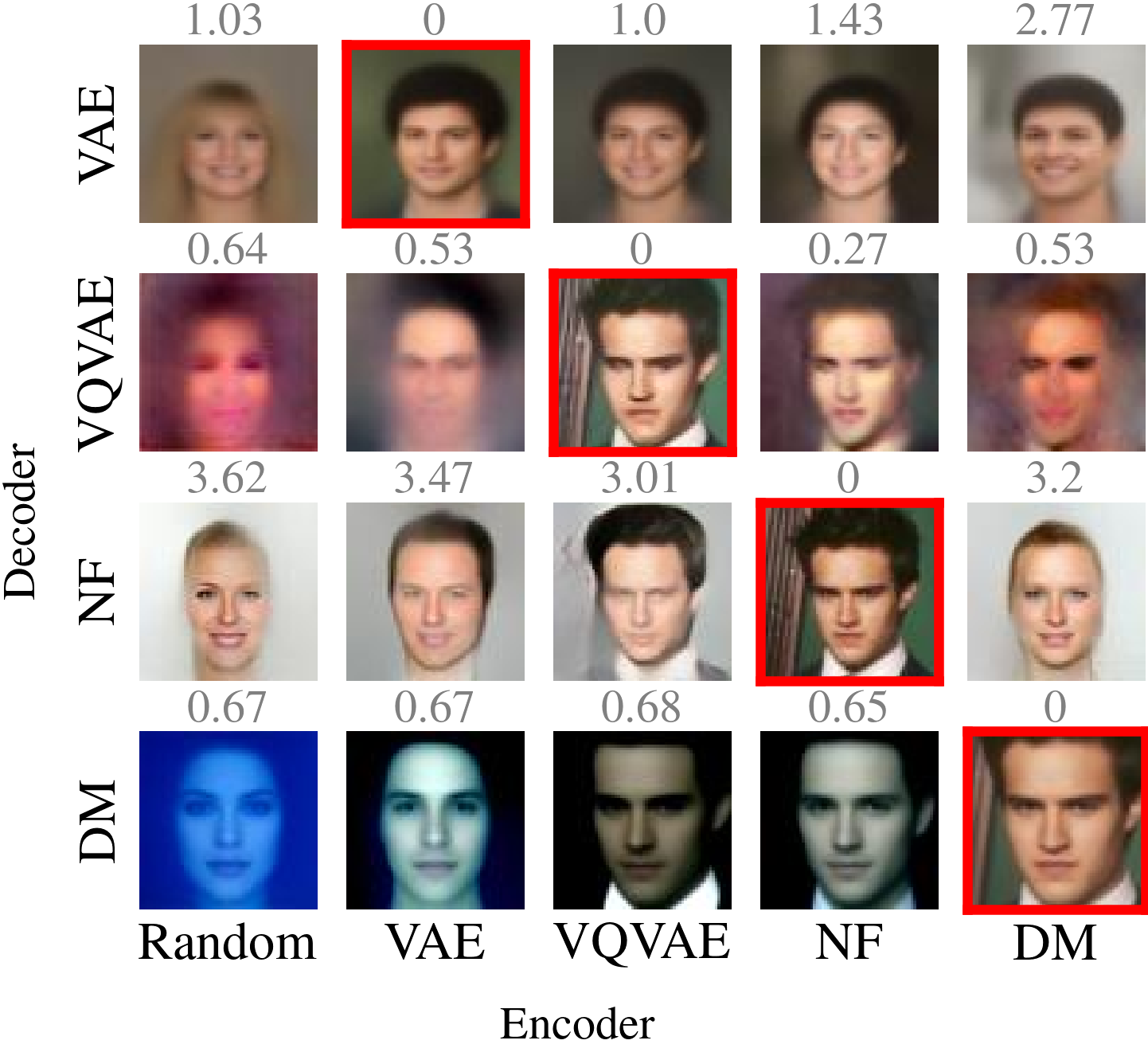}
\hspace{10mm}
 \raisebox{2\height}{
\includegraphics[width=0.07\columnwidth]{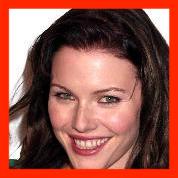}
}
\includegraphics[width=0.35\columnwidth]{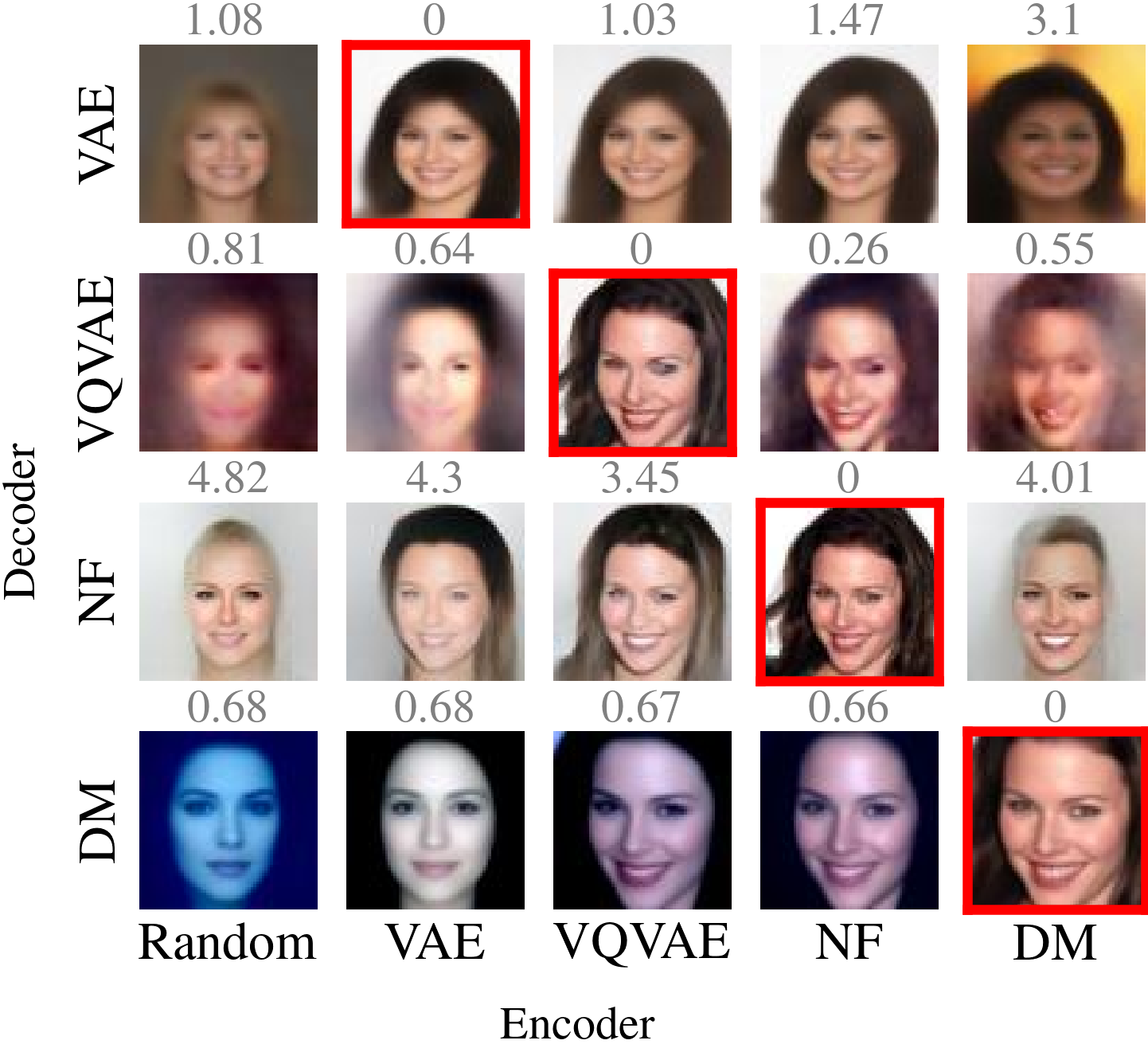}
\caption{See caption of Figure \ref{fig:celeba-mapped}.}
\end{figure}

\begin{figure}[h]
\centering
\includegraphics[width=0.4\columnwidth]{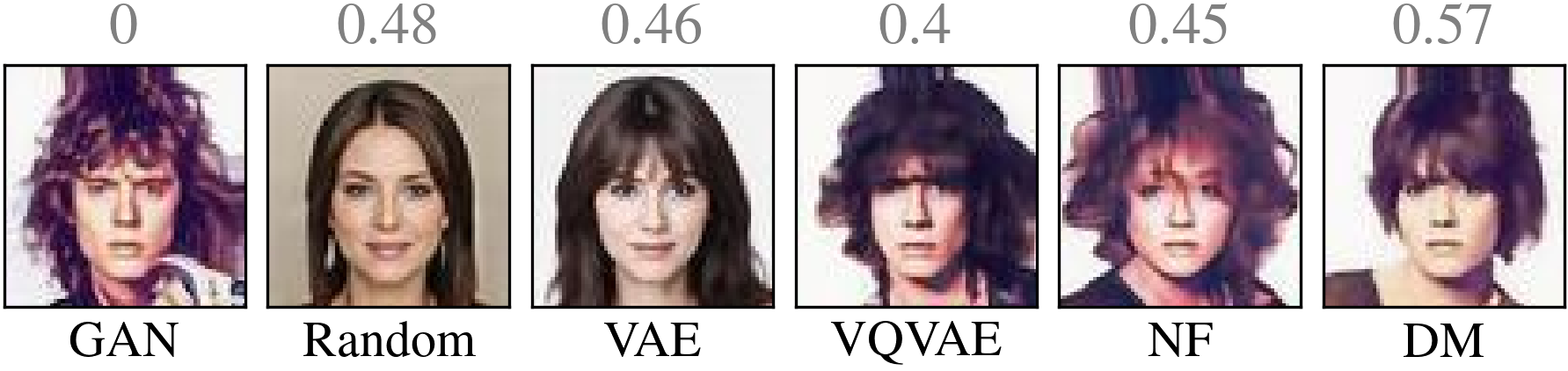}
\hspace{10mm}
\includegraphics[width=0.4\columnwidth]{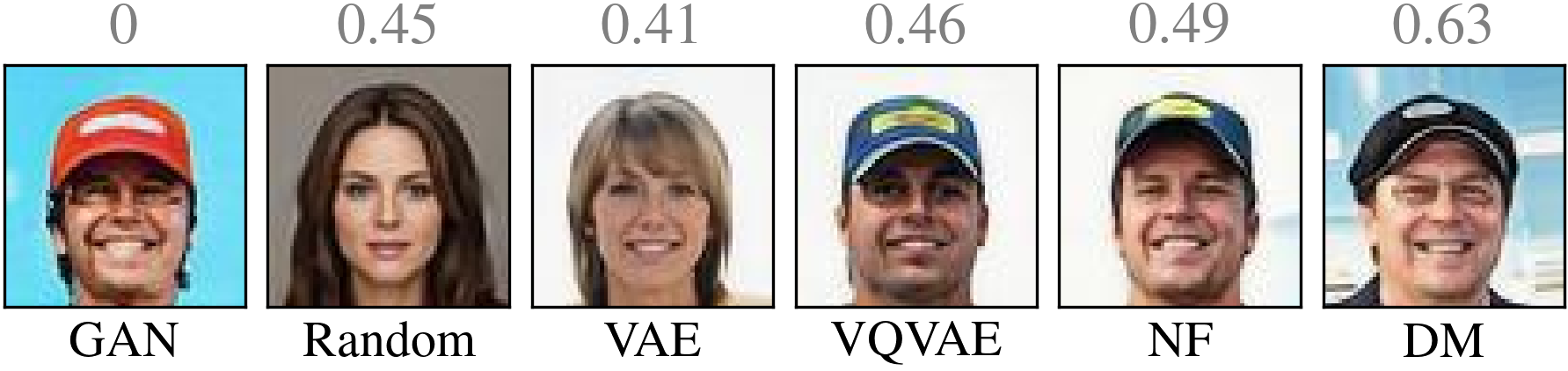}
\caption{See caption of Figure \ref{fig:celeba-mapped-gen}.}
\end{figure}

\end{document}